\documentclass[10pt,twocolumn,letterpaper]{article}

\usepackage{lineno}
\usepackage{cvpr}
\usepackage{times}
\usepackage{epsfig}
\usepackage{graphicx}
\usepackage{amsmath}
\usepackage{amssymb}
\usepackage{amsthm}
\usepackage{multirow}
\usepackage{amsmath}
\usepackage{url}
\usepackage{multicol}
\usepackage{booktabs}
\usepackage{algorithm}
\usepackage{algorithmic}
\usepackage{subfigure}
\usepackage{appendix}
\usepackage{wrapfig}
\usepackage{xcolor}
\usepackage{}
\cvprfinalcopy

  \newlength\titlebox 
\cvprfinalcopy 

\def\cvprPaperID{4129} 

\begin{document}

\title{A Main/Subsidiary Network Framework for Simplifying Binary Neural Networks}

\author {Yinghao Xu$^{1\ast}$\qquad Xin Dong$^{2}$*\qquad Yudian Li$^{4}$\qquad Hao Su$^{3}$
\\
\\
$^{1}$Zhejiang University \qquad
$^{2}$Harvard University \\
$^{3}$University of California, San Diego\\
$^{4}$University of Electronic Science and Technology of China \\
{\tt\small {$^{1}$justimyhxu@zju.edu.cn\qquad $^{2}$xindong@g.harvard.edu \qquad $^{4}$daniellee2519@gmail.com \qquad $^{3}$haosu@eng.ucsd.edu}}
}

\maketitle
\begin{abstract}
    
To reduce memory footprint and run-time latency, techniques such as neural network pruning and binarization have been explored separately. However, it is unclear how to combine the best of the two worlds to get extremely small and efficient models. In this paper, we, for the first time, define the filter-level pruning problem for binary neural networks, which cannot be solved by simply migrating existing structural pruning methods for full-precision models. A novel learning-based approach is proposed to prune filters in our main/subsidiary network framework, where the main network is responsible for learning representative features to optimize the prediction performance, and the subsidiary component works as a filter selector on the main network. To avoid gradient mismatch when training the subsidiary component, we propose a layer-wise and bottom-up scheme.

We also provide the theoretical and experimental comparison between our learning-based and greedy rule-based methods.  

Finally, we empirically demonstrate the effectiveness of our approach applied on several binary models, including binarized NIN, VGG-11, and ResNet-18, on various image classification datasets. 
For binary ResNet-18 on ImageNet, we use 78.6\% filters but can achieve slightly better test error 49.87\%~(50.02\%-0.15\%) than the original model.
\end{abstract}

\section{Introduction}

Deep neural networks (DNN), especially deep convolution neural networks (DCNN), have made remarkable strides during the last decade. From the first ImageNet Challenge winner network, AlexNet, to the more recent state-of-the-art, ResNet, we observe that DNNs are growing substantially deeper and more complex. These modern deep neural networks have millions of weights, rendering them both memory-intensive and computationally expensive. To reduce the computational cost, the research into network acceleration and compression emerges as an active field.  

A family of popular compression methods are the DNN pruning algorithms, which are not only efficient in both memory and speed, but also enjoy
relatively simple procedure and intuition. This line of research is motivated by the theoretical analysis and empirical discovery that redundancy does exist in both human brains and several deep models~\cite{de2017ultrastructural,denil2013predicting}. We can categorize existing researches according to the level of the object, such as connection (weights)-level pruning, unit/channel/filter-level pruning, and layer-level pruning~\cite{Wen2016}. Connection-level pruning is the most widely studied approach, which produces sparse networks whose weights are stored as sparse tensors. Although both the \emph{footprint memory} and the \emph{I/O consumption} are reduced~\cite{Han2015}, such methods are often not helpful towards the goal of \emph{computation acceleration} unless specifically-designed hardware is leveraged. This is because the dimensions of the weight tensor remain unchanged, though many entries are zeroed-out. As a well-known fact, the MAC operations on random structured sparse matrices are generally not too much faster than the dense ones of the same dimension. In contrast, \emph{structural pruning techniques}~\cite{Wen2016}, such as unit/channel/filter-level pruning, are more hardware-friendly, since they aim to produce tensors of reduced dimensions or having specific structures. 
Using these techniques, it is possible to achieve both computation acceleration and memory compression on general hardware and is common for deep learning frameworks. 

We consider the structural network pruning problem for a specific family of neural networks -- binary neural networks. A binary neural network is a compressed network of a general deep neural network through the \emph{quantization strategy}. Convolution operations in DCNN\footnote{Fully connected layers can be implemented as convolution. Therefore, in the rest of the paper, we mainly focus on convolutional layers.} inherently involve matrix multiplication and accumulation (MAC). MAC operations become much more energy efficient if we use low-precision (1 bit or more) fixed-point number to approximate weights and activation functions (i.e., to quantify neurons)~\cite{Rastegari2016}. To the extreme extent, the MAC operation can even be degenerated to Boolean operations, if both weights and activation are binarized. Such binary networks have been reported to achieve $\sim$58x computation saving and $\sim$32x memory saving in practice. However, the binarization operation often introduces noises into DNNs~\cite{Li2017}, thus the representation capacity of DNNs will be impacted significantly, especially when we also binarize the activation function. Consequently, binary neural networks inevitably require larger model size (more parameters) to compensate for the loss of representation capacity. 

Although Boolean operation in binary neural networks is already quite cheap, even smaller models are still highly desired for low-power embedded systems, like smart-phones and wearable devices in virtual reality applications. Even though quantization (e.g., binarization) has significantly reduced the redundancy of each weight/neuron representation, our experiment shows that there is still heavy redundancy in binary neural networks, in terms of network topology. In fact, quantization and pruning are orthogonal strategies to compress neural networks: 
Quantization reduces the precision of parameters such as weights and activations, while pruning trims the connections in neural networks so as to attain the tightest network topology. However, previous studies on network pruning are all designed for full-precision models and cannot be directly applied for binary neural networks whose both weights and activations are 1-bit numbers. For example, it no longer makes any sense to prune filters by comparing the magnitude or $L_1$ norm of binary weights, and it is nonsensical to minimize the distance between two binary output tensors.

\emph{We, for the first time, define the problem of simplifying binary neural networks and try to learn extremely efficient deep learning models by combining pruning and quantization strategies.} Our experimental results demonstrate that filters in binary neural networks are redundant and learning-based pruning filter selection is constantly better than those existing rule-based greedy pruning criteria (by weight magnitude or $L_1$ norm).

We propose a learning-based method to simplify binary neural network with a main-subsidiary framework, where the main network is responsible for learning representative features to optimize the prediction performance, while the subsidiary component works as a filter selector on the main network to optimize the efficiency. The contributions of this paper are summarized as follows:

\begin{itemize}
    \item We propose a learning-based structural pruning method for binary neural networks to significantly reduce the number of filters/channels but still preserve the prediction performance on large-scale problems like the ImageNet Challenge. 
    \item We show that our non-greedy learning-based method is superior to the classical rule-based methods in selecting which objects to prune. We design a main-subsidiary framework to iteratively learn and prune feature maps. Limitations of the rule-based methods and advantages of the learning-based methods are demonstrated by theoretical and experimental results. In addition, we also provide a mathematical analysis for $L_1$-norm based methods. 
    \item To avoid gradient mismatch of the subsidiary component, we train this network in a layer-wise and bottom-up scheme. Experimentally, the iterative training scheme helps the main network to adopt the pruning of previous layers and find a better local optimal point. 
\end{itemize}

 \begin{figure*}[h]
        \centering
        \includegraphics[width=4in]{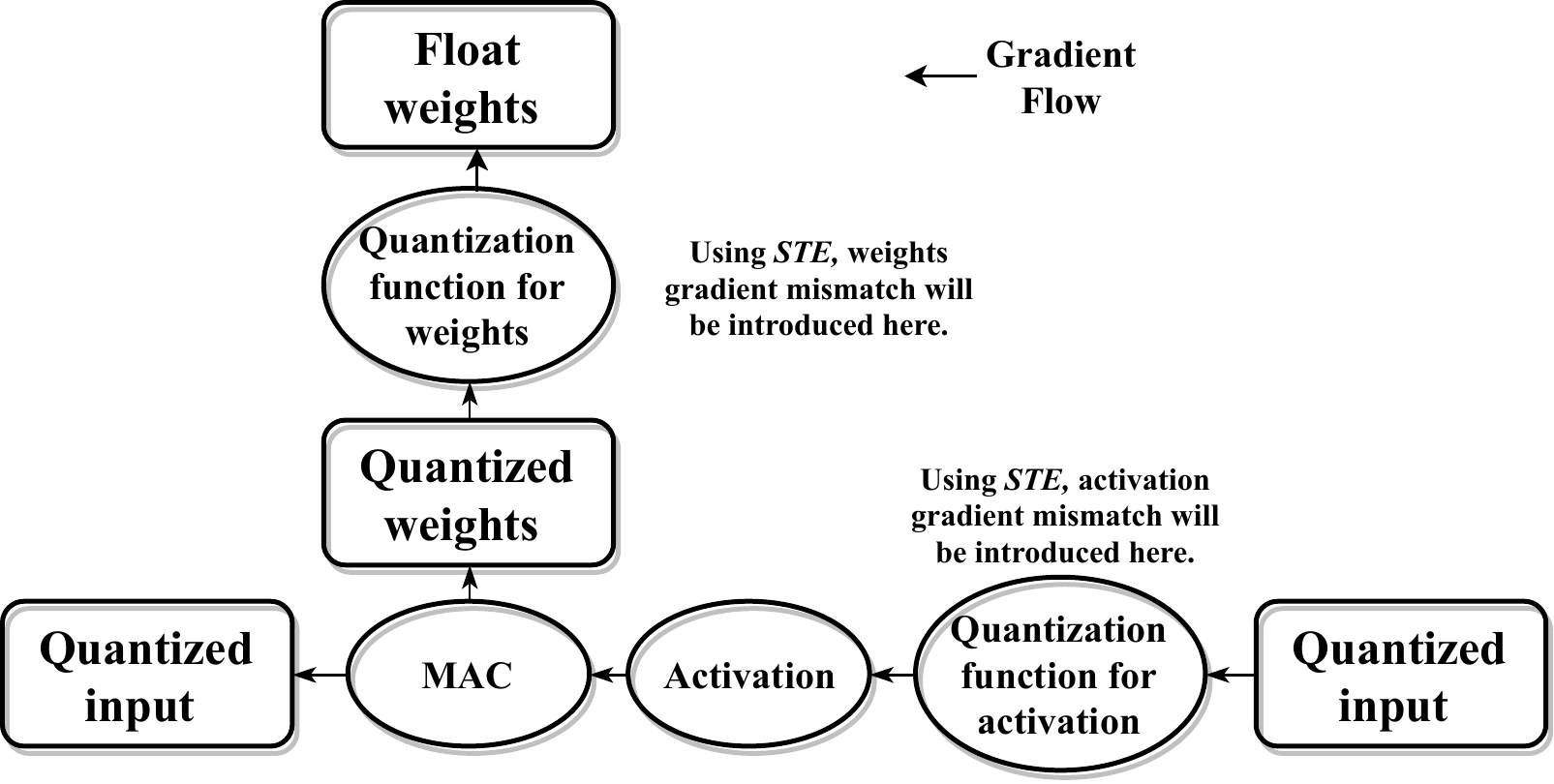}
        \caption{Gradient flow of binary neural networks during back-propagation. Rectangles represent the weight tensor and ellipses represent functional operation. In this paper, we use binary operation as a special quantization function. MAC is short for multiplication and accumulate operations, or the equivalent substitution like XNOR~\cite{Rastegari2016} in BNN.}
        \label{fig:my_label}
        \end{figure*}
        
\section{Related Work}
\subsection{Pruning}
    Deep Neural Network pruning has been explored in many different ways for a long time. \cite{Hassibi:1993:OBS:2987189.2987223} proposed Optimal Brain Surgeon (OBS) to measure the weight importance using the second-order derivative information of loss function by Taylor expansion. \cite{dong2017learning} further adapts OBS for deep neural networks and has reduced the retraining time.
    Deep Compression~\cite{Han2015} prunes connections based on weight magnitude and can achieve great compression ratio. The idea of dynamic masks~\cite{Guo2016} is also used for pruning. Other approaches used Bayesian methods and exploited the diversity of neurons to remove weights~\cite{Molchanov2017,mariet2016diversity}. However, these methods focus on pruning independent connection without considering group information. Even though they harvest sparse connections, it is still hard to attain the desired speedup on hardware.
    
    To address the issues in connection-level pruning,  researchers proposed to increase the group-sparsity by applying sparse constraints to the channels, filters, and even
      layers~\cite{Wen2016,alvarez2016learning,polyak2015channel,anwar2017structured}.  \cite{he2017channel} used LASSO constraints and reconstruction loss to guide network channel selection.  \cite{li2016pruning} introduced $L_1$-Norm rank to prune filters, which reduces redundancy and preserves the relatively important filters using a greedy policy. \cite{liu2017learning} leverages a scaling factor from batch normalization to pruning channels. To encourage the scaling factor to be sparse, a regularization term is added to the loss function. On one hand, methods mentioned above are all designed for full-precision models and cannot be trivially transferred to binary networks. For example, to avoid introducing any non-Boolean operations, batch normalization in binary neural networks~(like XNOR-Net) typically doesn't have scaling ($\gamma$) and shifting ($\beta$) parameters~\cite{Rastegari2016}. Since  all weights and activation only have two possible values $\{1,-1\}$, it is also invalid to apply classical tricks such as ranking filters by their $L_1$-Norms, adding a LASSO constraint, or minimizing the reconstruction error between two binary vectors. On the other hand, greedy policies that ignore the correlations between filters cannot preserve all important filters.

\subsection{Quantization}
Recent work shows that full precision computation is not necessary for the training and inference of DNNs~\cite{gupta2015deep}. Weights quantization is thus widely investigated, e.g., to explore 16-bit~\cite{gupta2015deep} and 8-bit~\cite{dettmers20158} fixed-point numbers. To achieve higher compression and acceleration ratio, extremely low-bit models like binary weights~\cite{courbariaux2015binaryconnect,hu2018hashing} and ternary weights\cite{zhu2016trained,zhou2017incremental,wang2017fixed} have been studied, which can remove all the multiplication operations during computation. Weight quantization has a relatively milder gradient mismatch issue as analyzed in Section~\ref{mismatch}, and lots of methods can achieve comparable accuracy with full-precision counterparts on even large-scale tasks. However, the ultimate goal for quantization networks is to replace all MAC operations by Boolean operations, which naturally desires that both activation and weights are quantized, even binarized. 

The activation function of quantized network has the form of a step function, which is discontinuous and non-differentiable. Gradient cannot flow through a quantized activation function during back-propagation. The \textit{straight-through estimator} (STE) is widely adopted to circumvent this problem, approximating the gradient of step function as $1$ in a certain range~\cite{hinton2012neural,bengio2013estimating}. \cite{Cai2017} proposed the Half-wave Gaussian Quantization (HWGQ) to further reduce the mismatch between the forward quantized activation function and the backward ReLU~\cite{nair2010rectified}. Binary Neural Networks (BNN) proposed in \cite{Courbariaux} and \cite{Rastegari2016} use only 1 bit for both activation functions and weights, ending up with an extremely smaller and faster network. BNNs inherit the drawback of acceleration via quantization strategy and their accuracy also needs further improving.

\section{Approach}
\label{headings}
Let $F_{b}^{i} \in R^{N_i \times H_i \times W_i}$ denote binary input feature maps of the $i$-th layer in an $I$-layer binary neural network, where $N_i$, $H_i$, and $W_i$ are the number of the input feature maps, height, and width of the activation map, respectively. Kernel weights $W_{b}^{i} \in R^{N_{i+1} \times N_{i} \times K_{i+1} \times K_{i+1}}$ in this layer are convolved with the input feature map $F_{b}^{i}$ into output feature map $F_{b}^{i+1}$. Because both weights and activations are binary, we remove the subscripts of $F_{b}$ and $W_{b}$ for clarity. The goal of pruning is to remove certain filters $W_{n,:,:,:}^{i}$, $n\in \Omega$, where $\Omega$ is the indices of pruned filters. If a filter is removed, the corresponding output feature map of this layer (which is also the input feature map of next layer) will be removed, too. Furthermore, the input channels of all filters in the next layer would become unnecessary. If all filters in one layer can be removed, the filter-level pruning will upgrade to layer-level pruning naturally. The goal of our method is to remove as many filters as possible for binary neural networks which are already compact and have inferior numerical properties, thus this task is more challenging compared with pruning a full-precision model.

\begin{figure*}[h]
    \centering
    \includegraphics[width=5.5in]{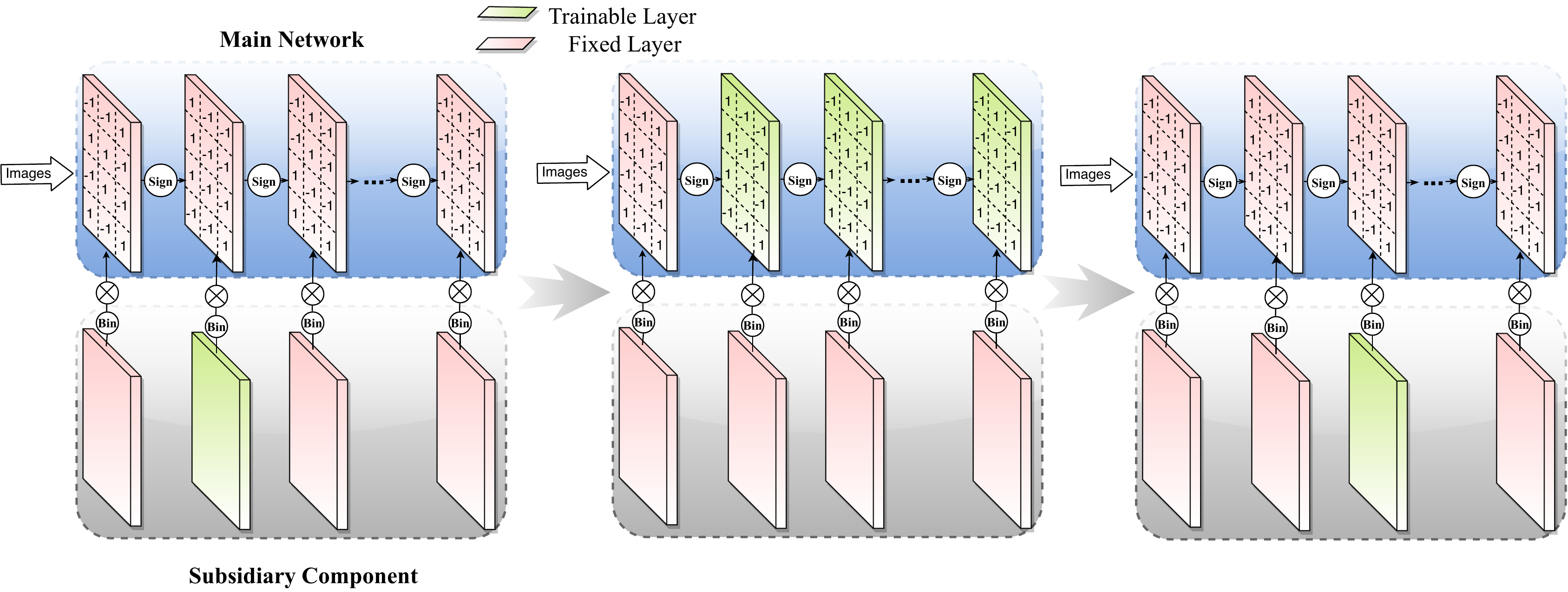}\vspace{-1.5mm}\caption{Pipline of our method. The main network in this figure is already pre-trained. From left to right: our subsidiary component training for i-th layer, main network retraining, and subsidiary component training for (i+1)-th layer.}
    \label{fig:my_label}
\end{figure*}

\vspace{-1mm}
\subsection{Subsidiary Component And main Networks}
We borrow the ideas from binary network optimization to simplify binary networks. While it sounds tautological, note that the optimization techniques were originally invented to solve the quantization problem, but we will show that it can be crafted to solve the pruning problem for binary networks. A new binary network, called subsidiary component, acts as learnable masks to screen out redundant features in the main network, which is the network to complete classification tasks. Each update of the subsidiary component can be viewed as the exploration in the mask search space. We try to find a (local) optimal mask in that space with the help of the subsidiary component.

The process of training subsidiary and main networks is as follows:

    \subsubsection{Feature Learning -- the Main Network}
    
    For layer $i$, the weights of subsidiary component $M^i \in R^{N_{i+1} \times N_{i} \times K_{i+1} \times K_{i+1}}$ are initialized by the uniform distribution: $M^i = \textit{U}(-\sigma,\sigma)$.
    In practice, $\sigma$ is chosen to be less than $10^{-5}$. To achieve the goal of pruning filters, all elements whose first index is the same share the same value. $M_{n,o_1,p_1,q_1}=M_{n,o_2,p_2,q_2}, \ \ \forall o,p,q$. Filter mask $O^i \in R^{N_{i+1} \times N_{i} \times K_{i+1} \times K_{i+1}}$ is an output tensor from the subsidiary component. In the first stage, we use the $Iden(\cdot)$ function~(identity transformation) to get $O^i$. 
    \[O^i = Iden(M^i)\]
    We apply the filter mask $O^i$ to screen main network's weights $W^i$,
    \[\hat{W}^i = O^i\otimes W^i \]
    , where $\otimes$ is element-wise product. $\hat{W}^{i}$ denotes the weights of the main network after transformation, which is used to be convolved with the input feature maps, $F^{i}$, to produce the output feature maps $F^{i+1}$.
    Then, weights of the main network, $W^j,\ j\in[1,I]$, are set to be \textbf{trainable} while weights of the subsidiary component, $M^j,\ j\in[1,I]$, are \textbf{fixed}. Because subsidiary weights are fixed and initialized to be near-zero, it will not function in the \textit{Feature Learning} stage, thus $\hat{W^j}\approx W^j,\ j\in[1,I]$.
    The whole main binary neural network will be trained from scratch.
    
    \subsubsection{Feature Selection -- the subsidiary component}
    \textbf{Training Subsidiary Component within a Single Layer $i$: }\label{onelayer}
    After training the whole main network from scratch, we use a binary operator to select features in a layer-wise manner. In opposite to the previous \textit{Feature Learning} stage, the weights of all layers $W^j,\ j\in[1,I]$ of the main network and the weights except layer $i$ of the subsidiary component $M^j,\ j\in[1,I]/[i]$ are set to be \textbf{fixed}, while the subsidiary component's weights at the current layer $M^i$ are \textbf{trainable} when selecting features for Layer $i$. The transformation function for the filter mask $O^i$ is changed from $Iden(\cdot)$ to $Bin(\cdot)$~(\textit{sign transformation}
    + \textit{linear affine}),
    \[O^i = Bin(M^i)=\frac{Sign(M^i)+1}{2} \]
    By doing this, we project the float-point $M_i$ to binarized numbers ranging from $0$ to $1$.
    Elements in $O_i$ which are equal to $0$ indicate that the corresponding filters are removed and the elements of value $1$ imply to keep this filter. 
    
    Since $Bin(\cdot)$ is not differentiable, we use the following function instead of the sign function in back propagation when training the subsidiary component $M^i$~\cite{hinton2012neural,bengio2013estimating},
    \begin{equation}
    \label{backward}
    f(x)=
    \begin{cases}
    x& {-1<x<1}\\
    1& {x\ge1}\\
    -1& {x\le-1}
    \end{cases}
    \end{equation}
    Apart from the transformation, we also need to add regularization terms to prevent all $O_i$ from degenerating to zero, which is a trivial solution. So the loss function of training Layer $i$ in the subsidiary component is,
    \begin{equation}
        \label{loss}
        \mbox{arg} \min_{M^i}\ \  L_{cross\_entropy} + \alpha \cdot L_{reg} + \beta\cdot L_{distill}
    \end{equation}
    \[L_{reg} = { \left\| { O }_{ i } \right\|  }_{ 1 }\]
    where $L_{cross\_entropy}$ is the loss on data and $L_{distill}$ is the distillation loss defined in (\ref{distill}).
    
    Finally, we fix the layers $M^j,\ j\in[1,I]$ in the subsidiary component and layers before $i$ in the main network (i.e., $W^j,\ j\in[1,i-1]$), and retrain the main layers after Layer $i$ (i.e., $W^j,\ j\in[i,I]$). 
    
    \textbf{Bottom-up Layer-wise Training for Multiple Layers:}
    \label{mismatch}
    We showed how to train a layer in the subsidiary component above. To alleviate the gradient mismatch and keep away from the trivial solution during \textit{Features Selection}, next, we propose a layer-wise and bottom-up training scheme for the subsidiary component: Layers closer to the input in the subsidiary component will be trained with priority. As Layer $i$ is under training, all previous layers (which should have already been trained) will be fixed and subsequent layers will constantly be the initial near-zero value during training. There are three advantages of this training scheme.
    
    First, as in~(\ref{backward}), we use \textit{STE} as in \cite{hinton2012neural,bengio2013estimating} to approximate the gradient of the sign function. By \textit{chain rule}, for each activation node $j$ in Layer $i$, we would like to compute an ``error term'' $\delta_j^i=\frac{\partial{L}}{\partial{a^i_j}}$ which measures how much that node is responsible for any errors in the output. For binary neural networks, activation is also binarized by a sign function which need \text{STE} for back-propagation. The ``Error term'' for binary neural networks is given by,
    \begin{multicols}{2}
    \begin{equation}
    \label{chainrule}
        \delta_j^i = Sign'(a_j^i)\cdot\sum_qw_{j,q}^{i+1}\delta^{i+1}_q
    \end{equation}
    \begin{equation}
    \label{ste}
        \frac{\partial{Sign(a_j^i)}}{\partial{a_j^i}} = \mathrm{1}_{|a^i_j| \leq 1}
    \end{equation}
    \begin{equation}
    \label{wgrad}
        \frac{\partial{L}}{\partial{M^i}} = \frac{\partial{L}}{\partial{O^i}}\cdot \frac{\partial{O^i}}{\partial{M^i}}
    \end{equation}
       \begin{equation}
    \label{steactivation}
        \frac{\partial{O^i}}{\partial{M^i}} = \frac{1}{2}\cdot\displaystyle 1_\mathrm{|M^i|\leq 1}
    \end{equation}
    \end{multicols}
    where (\ref{chainrule}) and (\ref{wgrad}) can be obtained by the \textit{chain rule}, and (\ref{ste}) and (\ref{steactivation}) are estimated from \textit{STE}, which will introduce gradient mismatch into back-propagation. We refer (\ref{steactivation}) as weight gradient mismatch issue and (\ref{ste}) as activation gradient mismatch issue. They are two open problems in the optimization of binary neural networks, both caused by the quantization transform functions like $Sign(\cdot)$. Starting from bottom layers, we can train and fix layers who are harder to train as early as possible for the subsidiary component. In addition, because of the retraining part in \textit{Features Selection}, bottom-up training scheme allows bottom layers to be fixed earlier, as well. In practice, this scheme results in more stable training curves and can find a better local optimal point.
    
    Second, the bottom-up layer-wise training scheme helps the main network to better accommodate the feature distribution shift caused by the pruning of previous layers. As mentioned before, the main difference in the motivation between our pruning method and rule-based methods is that we have more learnable parameters to fit the data by focusing on the final network output. With the bottom-up and layer-wise scheme, even if the output of Layer $i$ changes, subsequent layers in the main network can accommodate this change by modifying their features. 
    
    Lastly and most importantly, we achieve higher pruning ratio by this scheme. According to our experiments, a straight-forward global training scheme leads to limited pruning ratio. Some layers are pruned excessively and hence damaged the accuracy, while some layers are barely pruned, which hurts the pruning ratio. The layer-wise scheme would enforce all layer to be out of the comfort zone and allow balancing between accuracy and pruning ratio.

    \subsubsection{Pipeline}
    The pipeline of our method is as follows:
    
    \begin{enumerate}
        \item Initialize weights of subsidiary component $M^j,\ j\in[1,I]$ with near-zero $\sigma$'s. 
        \item Set $M^j,\ j\in[1,I]$ to be fixed, and train the whole main network from scratch.
        
        \item Train starting from the first binary kernel. Each layer is the same as in the algorithm shown below:
        \begin{itemize}
            \item Change the activation function for $M^{i}$ from $Iden(\cdot)$ to $Bin(\cdot)$. And all other parameters apart from $M^{i}$ are fixed. Train subsidiary component according to (\ref{loss}).
            \item Fix the subsidiary layers $M^{j}$, $j
            \in [1,I]$ and main layers before i-th layer $W^{j},\ j \in [1,i-1]$, and retrain main layers after i-th layer $W^{j},\ j\in[i,I]$.
        \end{itemize}
    
    \end{enumerate}

    \subsubsection{Distillation loss}
\begin{figure}[h]
     \includegraphics[width=7cm]{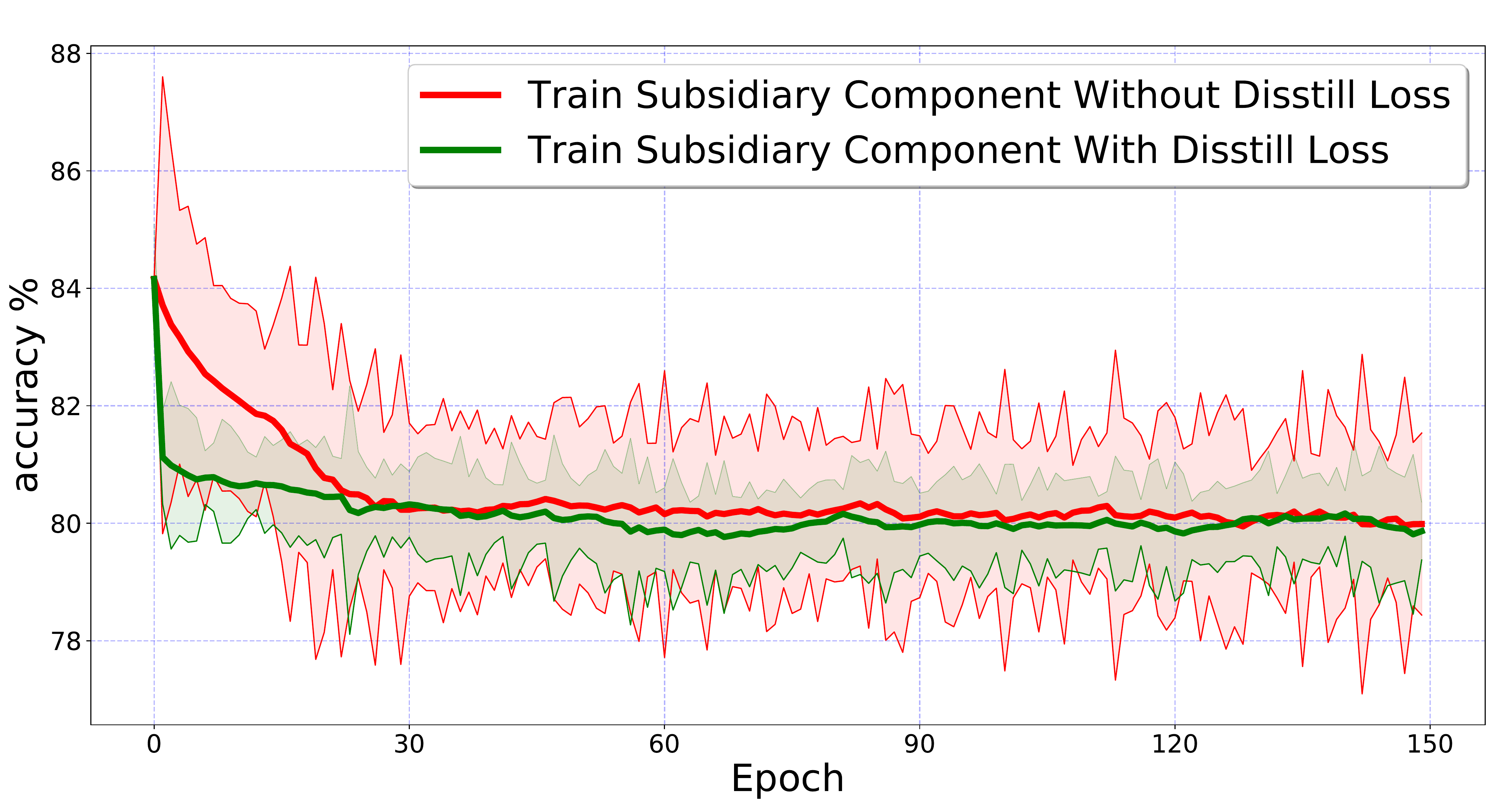}
    \caption{The learning curve for the Subsidiary Component. The red line refers to the learning curve without the distillation loss, and the red background represents the learning curve variance of every epoch. The green line and background represent the subsidiary component learning curve and the variance of each epoch. Clearly, the distillation loss makes the training procedure more stable.}
    \label{fig:f3}  
    \end{figure}

    Though pruning network filters is not an explicit transfer learning task, the aim is to guide the thin network to learn more similar output distributions with the original network. The model is supposed to learn a soft distribution but not a hard one as proposed in previous traditional classifier networks. Hence, we add a distillation loss to guide the training subsidiary component to be more stable, as shown in Figure \ref{fig:f3}.
    \begin{equation}
    \label{distill}
    L_{distill} = (p\|q) = H(p,q)- H(p)
    \end{equation} 
    We set $p$ to be the original binary neural network distribution. Because the distribution is fixed, the $H(p)$ is a constant and can be removed from $L_{distill}$.
    It means that the distillation loss can be written as 
    \[L_{distill} = -\sum_{i = 1}^{M} log\frac{exp(z_i/T)}{\sum_{j=1}^{M} exp(z_j/T) } \times \frac{exp(t_i/T)}{\sum_{j}^{M} exp(t_j/T)}\]
    where $z_i$ and $t_i$ represent the final output of the pruned and original networks before the softmax layer. $T$ is a temperature parameter for the distillation loss defined in \cite{hinton2015distilling}. We set $T$ as $1$ in practice. $M$ is the number of classes.

\subsection{Comparison with rule-based methods}
Previous methods use rules to rank the importance of each filter and then remove the top $k$ least important filters. The rules can be weight magnitude, e.g., measured by the $L_1$ norm, or some other well-designed criteria. 

Studies in this line share the same motivation that individual filters have their own importance indication, and filters with less importance can be removed relatively safely. This assumption ignores interactions among filters. As mentioned before, rule-based pruning algorithms use a greedy way to prune filters, i.e., they assume that individual filters behave independently and their own importance (or function) for representation learning. We give a theoretical analysis in Section~\ref{l1-norm} about this point. In fact, pruning filters independently may cause problems when filter are strongly correlated. For example, if two filters have learned the same features (or concepts), these two filters may be pruned out together by rule-based methods, because their rankings are very close. Clearly, pruning one of them is a better choice.

However, almost all these criteria are based on value statistics and are completely unsuitable for the binary scenario with only two discrete values. One possible pruning method is to exhaustively search the optimal pruning set, but this is NP-Hard and prohibitive for modern DNNs that have thousands of filters. 
Our method uses the subsidiary component to ``search'' the optimal solution. 
Our soft ``search'' strategy is gradient-based and batch-based compared to exhaustive search, and it is much more efficient.
\subsection{Relation to $L_1$-Norm pruning}
\label{l1-norm}
If our main network is full-precision, the $L_1$-Norm based pruning technique would be strongly relevant to our method, except that we target at optimizing the final output of the network, whereas the $L_1$-Norm based method greedily controls the perturbation of the feature map in the next layer.

Suppose that $W=[w_1; \ldots; w_n]$ is the original filter blocked by rows, $W'=[w'_1;\ldots; w'_n]$ is the pruned filter, and $x$ is the input feature map. Let $\Delta w_i \equiv w_i-w'_i$. Then, the $L_1$-Norm approach minimizes the upper bound of the following problem: $\max_{\|x\|_{\infty}<\tau} \|Wx-W'x\|$. To see this, note 

    
    
\begin{align}
     & \|Wx-W'x\| = \|\begin{bmatrix} w_1-w'_1\\ \cdots\\ w_n-w'_n  \end{bmatrix} x\|=\|\sum_i \Delta w_i x\|  \\
     & \le \sum_i \|\Delta w_i x\|\le \sum_i \|\Delta w\|_1\|x\|_{\infty}\le \sum_i \|\Delta w\|_1 \tau
\end{align}

To minimize $\sum_i \|\Delta w\|_1$ by zeroing-out a single row $w_i$, obviously, the solution is to select the one with the smallest $L_1$-Norm. 

However, note that this strategy cannot be trivially applied for binary networks, because the $L_1$-Norm for any filter that is a $\{-1,+1\}$ tensor of the same shape is always identical.

\subsection{Relation to LASSO regression based least reconstruction error pruning}
Previous work~\cite{He} uses the LASSO regression to minimize the reconstruction error of each layer: $min\ \|Y-\sum_{i=1}^L\beta_iX_iW_i\|_F^2,\ \ \|\beta\|_0\leq C'$. Solving this $L_0$ minimization problem is NP-hard, so the $L_0$ regularization is usually relaxed to $L_1$. In the binary/quantization scenario, activations only have two/several values and the least reconstruction error is not applicable. Instead of minimizing the reconstruction error of a layer, our method pays attention on the final network output with the help of the learnable subsidiary component. We directly optimize the discrete variables of masks (a.k.a subsidiary component) without the relaxation.

\section{Experiments}
    To evaluate our method, we conduct several pruning experiments for VGG-11, Net-In-Net~(NIN), and ResNet-18 on CIFAR-10 and ImageNet. Since our goal is to simplify binary neural networks, whose activation and weights are both 1-bit, all main models and training settings in our experiments inherit from XNOR-Net~\cite{rastegari2016xnor}. 
    Since we are, to the best of our knowledge, the first work to define filter-level pruning for binary neural networks, we proposed a rule-based method by ourselves as the baseline. Instead of ranking filters according to the $L_1$-Norm~\cite{li2016pruning}, we use the magnitude of each filter's scaling factor (MSF) as our pruning criterion. Inspired by \cite{li2016pruning}, we test both the ``prune once and retrain'' scheme\footnote{Prune filters of multiple layers at once and retrain them until the original accuracy is restored} and the ``prune and retrain iteratively'' scheme\footnote{Prune filters layer by layer and then retrain iteratively. The model is retrained before pruning the next layer for the weights to adapt to the changes from the pruning process.}.

    As pointed out in ~\cite{rastegari2016xnor} we set weights of the first layer and last layer as full-precision, which also means that we only do pruning for the intermediate binary layers. We measure effectiveness of pruning methods in terms of PFR, the ratio of the number of pruned filters to original filter number, and error rate before and after retraining. For error ratio, smaller is better. For PFR, larger is better. For CIFAR-10, when training the main network, learning rate starts from $10^{-4}$, and learning-rate-decay is equal to 0.1 for every 20 epochs. Learning rate is fixed with $10^{-3}$ when training the subsidiary component. For ImageNet, we set a constant learning rate of $10^{-3}$ for the subsidiary component and main work.

\begin{figure*}[h] \centering 
    \label{fig:pic}
    \hspace{-1mm}
    \subfigure[NIN, CIFAR-10] { \label{fig:a} 
    \includegraphics[width=0.46\columnwidth]{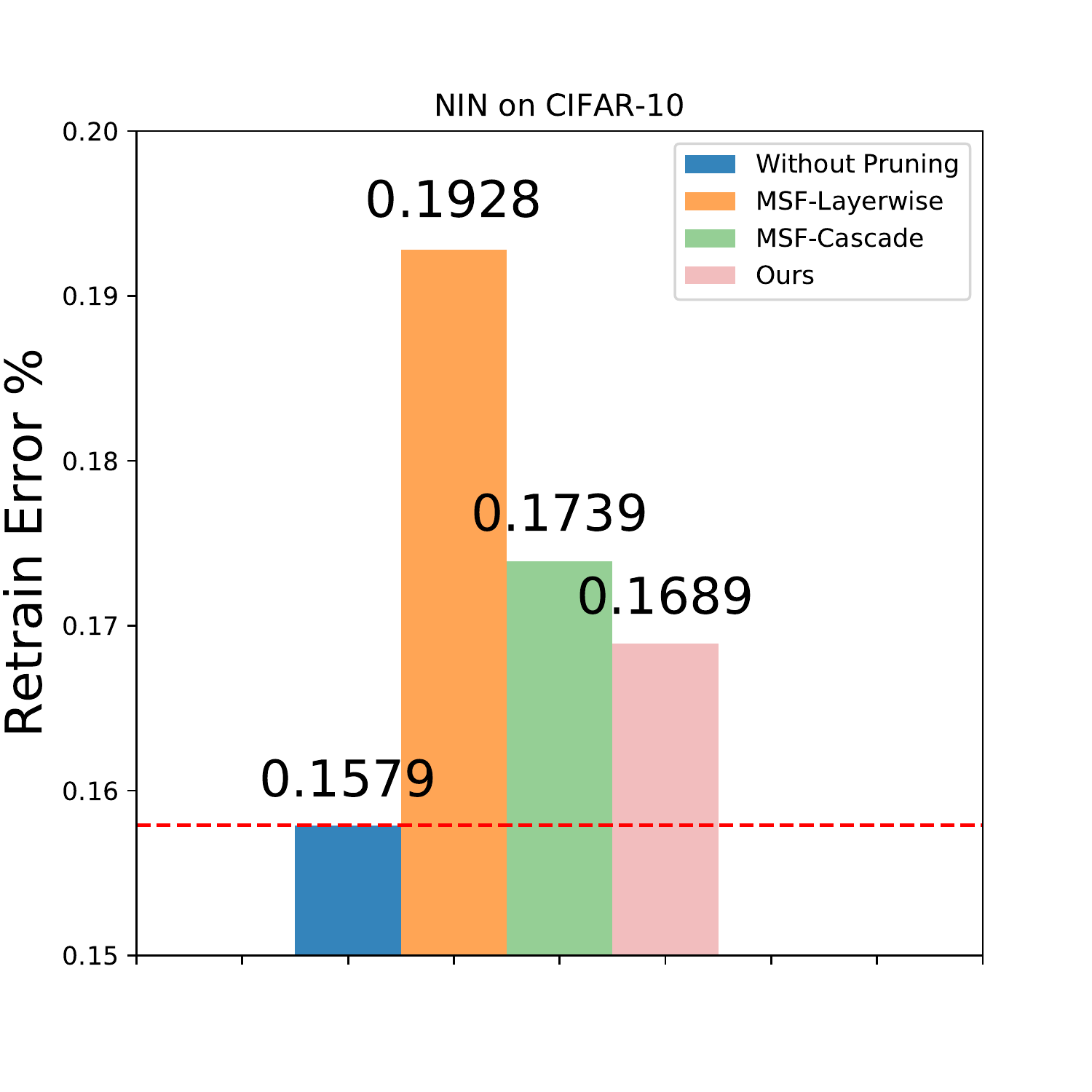} 
    } 
    \subfigure[VGG-11, CIFAR-10] { \label{fig:b} 
    \includegraphics[width=0.46\columnwidth]{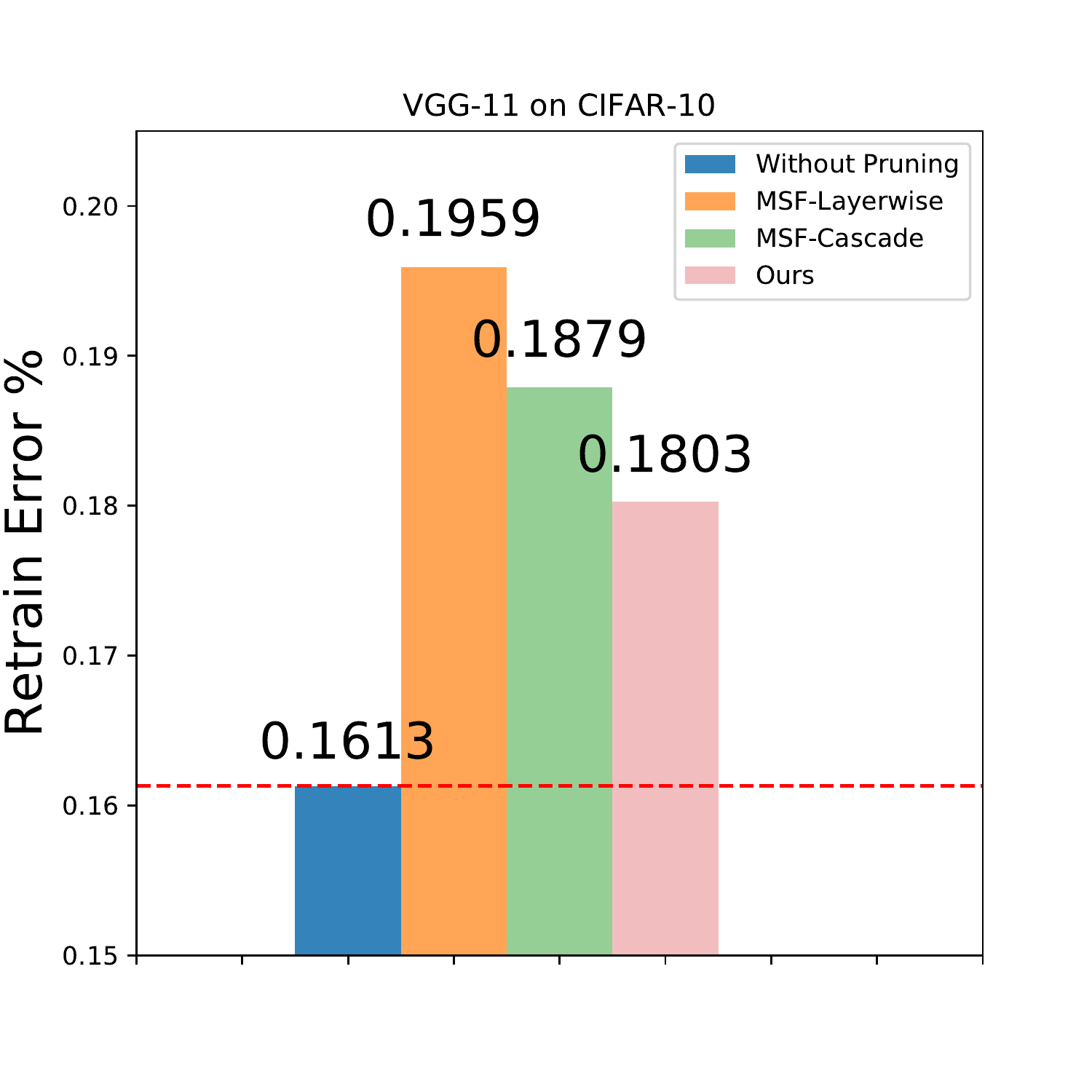} 
    } 
    \subfigure[ResNet-18, CIFAR-10] { \label{fig:a} 
    \includegraphics[width=0.46\columnwidth]{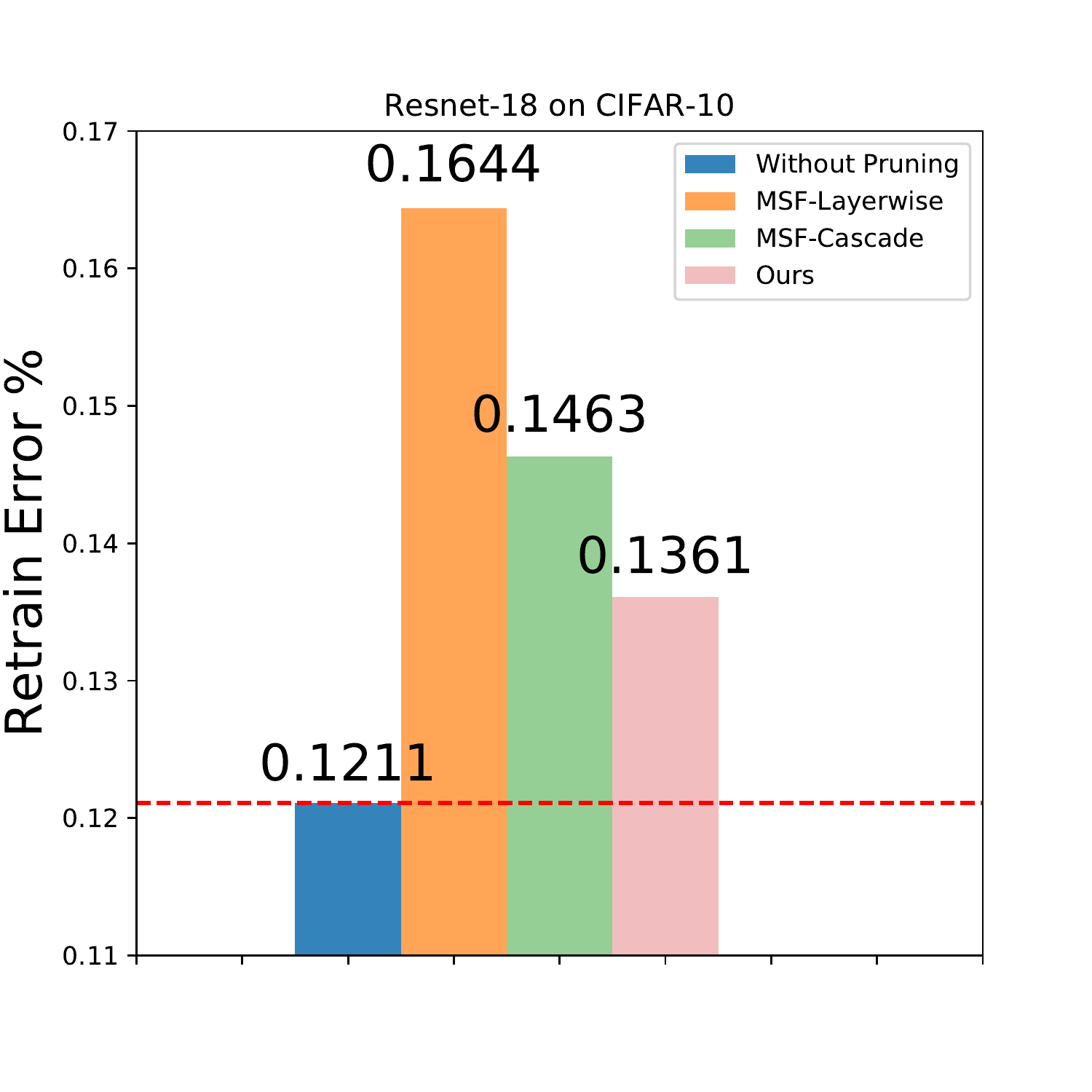} 
    } 
    \subfigure[ResNet-18, ImageNet ] { \label{fig:b} 
    \includegraphics[width=0.46\columnwidth]{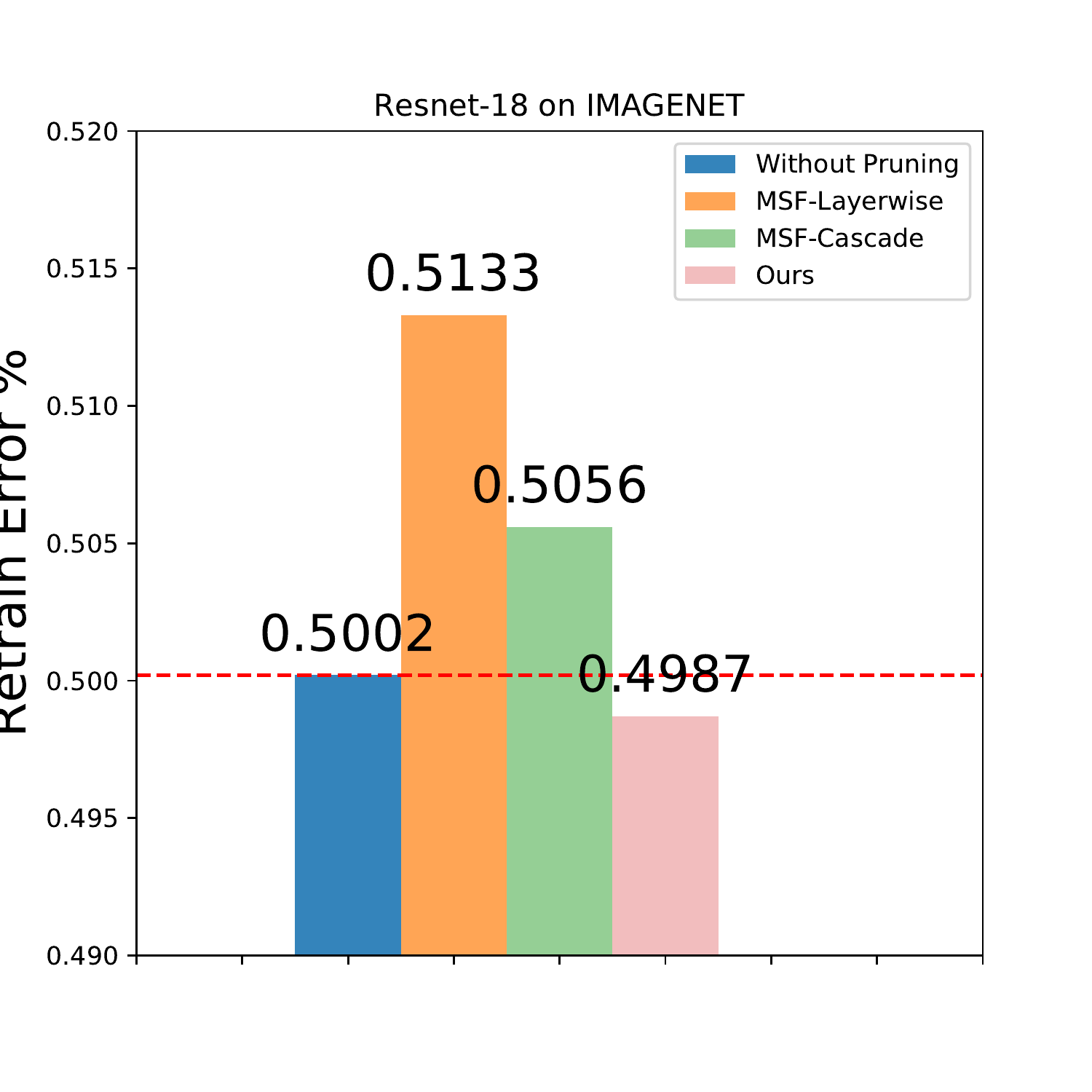} 
    } 
    \caption{Results. We control all methods to have the same PFR and compare their error after retraining. Compared with the baseline designed by us~(in orange and green), performance of our learning-based method~(in pink) is bette. Over the four plots, from left to right, the PFR is \textbf{33.05\%}, \textbf{39.70\%}, \textbf{39.89\%}, and 
    \textbf{21.40\%}. For ResNet-18 on ImageNet, we pruned 21.4\% of the filters while the retrained error decreased from 50.02\% to 49.87\% } 

    \end{figure*}

    For  fair comparison, we control PFR for each layer of these methods to be the  same to observe the final Retrain-Error. In Figure \ref{fig:pic}, MSF-Layerwise refers to the ``prune once and retrain'' scheme, and the MSF-Cascade refers the ``prune and retrain iteratively'' scheme. The first three figures of experiments were done on the CIFAR-10 dataset. The last figure  refers to results on Imagenet.

    \subsection{NIN and VGG-11 On CIFAR-10}
    
    \begin{table*}[h]
    \caption{Overall results}
    \vspace{2mm}
    \label{overall}
    \begin{center}
    \begin{tabular}{cccccccc}
    \multicolumn{1}{c}{\bf Method}  &\multicolumn{1}{c}{\bf Model} &\multicolumn{1}{c}{\bf Original Error(\%)}  &\multicolumn{1}{c}{\bf Retrain Error(\%)} & \multicolumn{1}{c}{\bf PFR(\%)} 
    \\ \toprule
    MSF-Layerwise     & NIN          & 15.79\%   & 19.28\%              &     33.05\%      \\
    MSF-Cascade       & NIN          & 15.79\%   & 17.39\%              &     33.05\%      \\
    Our Method              & NIN          & 15.79\%   & \textbf{16.89}\%     &     33.05\%      \\       \midrule
    MSF-Layerwise    & VGG-11       & 16.13\%   & 19.59\%              &     39.70\%      \\
    MSF-Cascade      &  VGG-11      & 16.13\%   & 18.79\%              &     39.70\%      \\  
    Our Method              & VGG-11       & 16.13\%   & \textbf{18.03}\%     &     39.70\%      \\       \midrule
    MSF-Layerwise     & ResNet-18    & 12.11\%   & 16.44\%              &     39.89\%      \\
    MSF-Cascade      &  ResNet-18   & 12.11\%   & 14.63\%              &     39.89\%      \\
    Our Method              &  ResNet-18   & 12.11\%   & \textbf{13.61}\%     &     39.89\%      \\       \midrule
    MSF-Layerwise     & ResNet-18    & 50.02\%   &51.33 \%              &     21.40\%      \\
 MSF-Cascade       &  ResNet-18   & 50.02\%   & 50.56\%&     21.40\%      \\
    Our Method              & ResNet-18    & 50.02\%   & \textbf{49.87}\%     &     21.40\%      \\       \bottomrule
    \end{tabular}
    \end{center}
    \end{table*}

   \begin{table*}[]
\begin{center}
\caption{FLOPs and Memory usage for our pruned model}
\vspace{2mm}

\begin{tabular}{llcccc}\toprule
                           &                        & FLOPs & Speedup & Memory Usage & Memory saving \\ \midrule
\multirow{3}{*}{ResNet-18} & Our Pruned Model       & 1.46 $\times$ $10^{8}$     &  12.39$\times$       &  30.87Mbit            &       12.11$\times$        \\
                           & XNOR-NET               & 1.67 $\times$ $10^{8}$       &  10.86$\times$       &  33.70Mbit            & 11.10$\times$              \\
                           & Full-precision Res-Net & 1.81 $\times$ $10^{9}$      &  ---       &    374.1Mbit          &        ----       \\  \bottomrule
\end{tabular}

\label{tab:mem}
\end{center}
\end{table*}
    NIN is a fully convolutional network, using two $1\times 1$ convolution layers instead of fully connected layer, and has quite compact architecture. VGG-11 is a high-capacity network for classification. VGG-11 on CIFAR-10 consists of 8 convolutional layers(including 7 binary layers) and 1 fully connected layer. 
    Batch normalization is used between every binary convolution and activation layer, which makes the training process more stable and converges with high performance. For both MSF-Layerwise and MSF-Cascade, with the same PFR, the performance is worse than us.  With 30\% $\sim$ 40\% of pruning filter ratio, the pruned network error rate only increased 1\% $\sim$  2\%.

    \subsubsection{Learning Rate is Important}
    
\begin{figure}[h]
    \centering
    \includegraphics[width=2.5in]{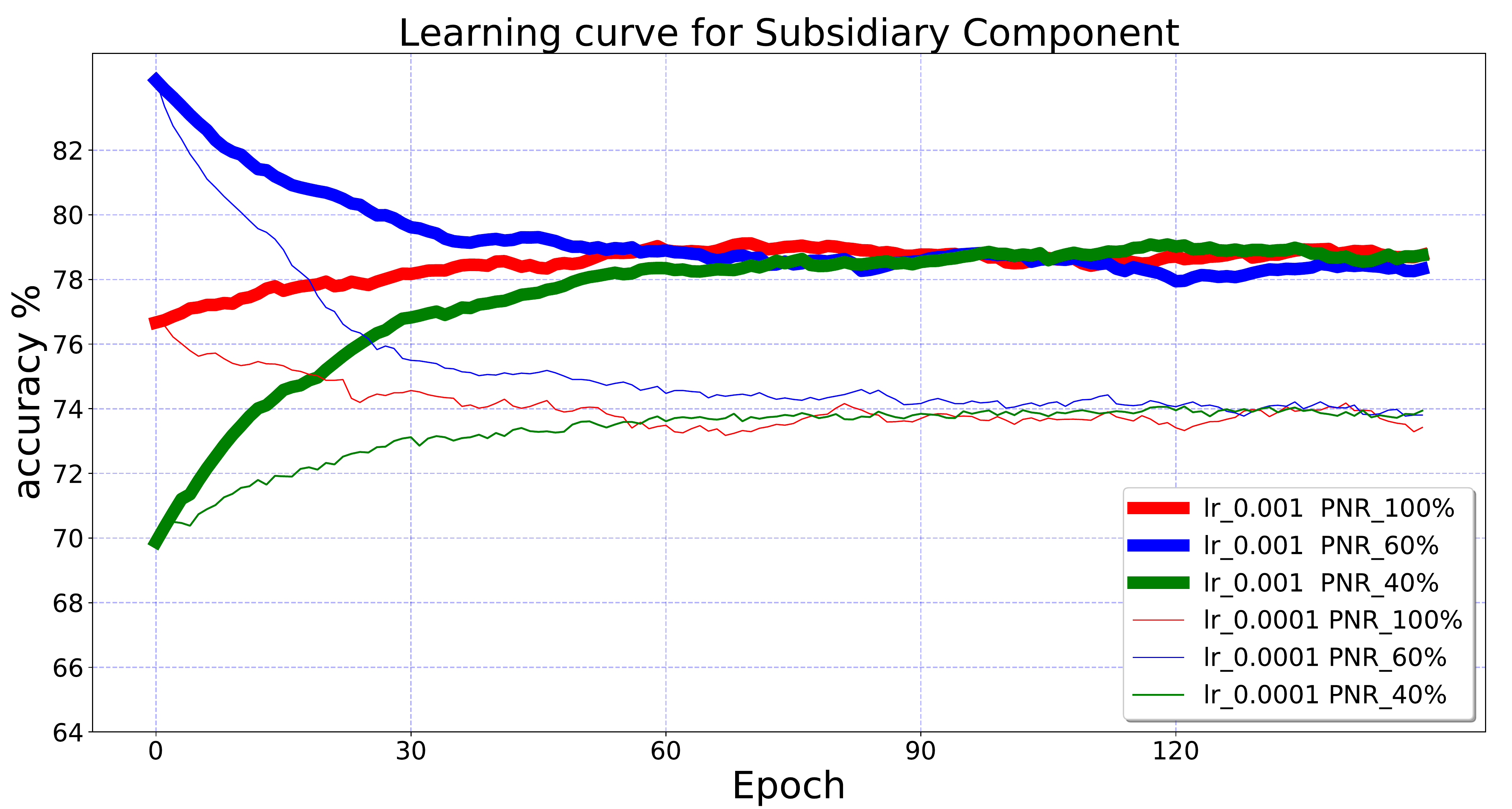}
   \caption{ Learning curve for subsidiary component. We train the subsidiary component with different learning rate. These curves are smoothed for  directly seeing the trend of the learning Subsidiary Component. All  dotted lines represent the learning curve of the large learning rate $10^{-3}$, the normal lines represent the learning curves of the small learning rate $10^{-4}$.} 
    \label{fig:lr}  
   
    \end{figure}

    An interesting phenomenon is observed when training subsidiary components for different models. 
    We try different learning rates in our experiments and observe it impacts final convergent point a lot as shown in Figure~\ref{fig:lr}.
    The relatively smaller learning rate~($10^{-4}$) will converge with lower accuracy and higher pruning number; however, the larger learning rate ~($10^{-3}$) leads to the opposite result. 
    
    One possible explanation is that the solution space of the high-dimensional manifold for binary neural networks is more discrete compared to full-precision networks, so it is difficult for a subsidiary component to jump out of a locally optimal point to a better one. Moreover, in the binary scenario, larger learning rate will increase the frequency of value changing for weights.
    Our motivation is to use a learnable subsidiary components to approximate exhaustive search, so using a larger learning rate will enable the subsidiary component to ``search'' more aggressively.  A large learning rate may be unsuitable for normal binary neural networks like the main network in this paper, but it is preferred by the subsidiary component.

\subsubsection{Initialization of subsidiary component is NOT Sensitive}
    As mentioned in section 3.1.1, we use the uniform distribution to initialize the mask. According to the expectation of the uniform distribution, $E (PNR) = 0.5$, where PNR is the ratio of the number of positive elements in subsidiary weights to size of weights.
    However, since we use $Sign(\cdot)$, different PNR may impact the result a lot.
    We conduct six experiments on different models across different layers and show that initialization with 0.4, 0.6, 1.0 PNR will all converge to the same state. 
     However, when PNR is 0.2, final performance will be very poor. A possible reason is that the number of filters thrown out by the initialization is too large, and due to the existence of the regularization term, the network's self-adjustment ability is limited and cannot converge to a good state. Hence we recommend the PNR to be intialized to greater than 0.4.

\subsection{ResNet on CIFAR-10 and ImageNet}
        Compared with NIN and VGG-11, ResNet has identity connections within residual block and much more layers. As the depth of network increases, the capacity of network also increases, which then leads to more redundancy. From experimental results, we find that when the identification mapping network has a downsampling layer, the overall sensitivity of the residual block will increase. Overall result for ResNet on CIFAR-10 is shown in table~(\ref{overall}), and statistics for each layer can be found in Appendix. 

        We further verify our method with ResNet-18 on ImageNet. $\alpha$ can be set from  $10^{-7}$ to $10^{-9}$ depending on the expected PFR, the accuracy and pruning ratio are balanced before retraining. 
        After 20 epoches retraining for each layer, the final PFR is 21.4\%, with the retrained error has decreased from $50.02\%$ to $49.87\%$.

\subsection{Efficiency and Memory Usage Analysis}
        In this section, we will analyze the speedup and memory saving of our pruned model and compare with XNOR-Net and full-precision network in ResNet-18. 
        
        The memory usage is the summation of number bits of all weights within one model. 
        In addition, we use FLOPs to measure the efficiency for our pruned model.  Because of the binary operation can implemented in XNOR operation and bit-counting in 64 parallel. So the final FLOPS are composed of the full-precision multiplication plus 1/64 1-bit multiplication.
        
        We keep the first convolution layer and the last fully-connected layer to be real-valued and keep other weights and activations in the whole network are all binarized. As shown in Table~\ref{tab:mem}, our pruned ResNet-18 model for ImageNet speed up 12.39$\times$ and reduces 12.11$\times$ memory usage, compared with the full-precision ResNet-18 Model. We also achieves up to 14.10\% higher speedup ratio and 9.09\% memory saving ratio compared with XNOR-Net, saying that our model requires less memory and fewer FLOPs.


\section{Conclusion}
        In this paper, we, for the first time, define the filter-level pruning problem for binary neural networks and propose a novel learning-based main/subsidiary network framework. Extensive experimental results on CIFAR and ImageNet demonstrate that the proposed main/subsidiary network framework and the novel training methods show efficiency for pruning of binary neural networks. What's more, our method is also friendly to pruning problem for quantized networks. 
        In the future, we will explore more advanced learning algorithms for subsidiary part of the framework, because the learning-based framework for pruning has important value and will be treated as future work either.

\newpage
{\small
\bibliographystyle{ieee}
\bibliography{egbib}
}

\end{document}